%% file: neurips_2026.tex
\documentclass[preprint]{article}
\PassOptionsToPackage{numbers,sort&compress}{natbib}

\usepackage{neurips_2026}

\usepackage[utf8]{inputenc} 
\usepackage[T1]{fontenc}    
\usepackage{hyperref}       
\usepackage{url}            
\usepackage{booktabs}       
\usepackage{amsfonts}       
\usepackage{nicefrac}       
\usepackage{microtype}      
\usepackage{xcolor}         
\usepackage{amsmath}
\usepackage{tikz}
\usetikzlibrary{arrows.meta, positioning}
\usepackage{algorithm}
\usepackage{algorithmic}

\title{Separating Intrinsic Ambiguity from Estimation Uncertainty in Deep Generative Models for Linear Inverse Problems}

\author{
Yuxin Guo$^{1,2}$\thanks{Corresponding author: yuxinguo@andrew.cmu.edu},
Dongrui Deng$^{1,2}$,
Pulkit Grover$^{1,3}$\\
$^1$Neuroscience Institute, Carnegie Mellon University\\
$^2$Machine Learning Department, Carnegie Mellon University\\
$^3$Department of Electrical and Computer Engineering, Carnegie Mellon University\\
}
\begin{document}

\maketitle

\begin{abstract}
Recently, deep generative models have been used for posterior inference in inverse problems, including high-stakes applications in medical imaging and scientific discovery, where the uncertainty of a prediction can matter as much as the prediction itself. However, posterior uncertainty is difficult to interpret because it can mix ambiguity inherent to the forward operator with uncertainty propagated through inference. We introduce a structural decomposition of posterior uncertainty that isolates intrinsic ambiguity. A cascade formulation makes this ambiguity accessible for calibration analysis, enabling qualitative diagnostics and simulation-based calibration tests that reveal failure modes that remain hidden when models are selected by reconstruction quality alone. We first validate the approach on a Gaussian example with analytical posterior structure, then illustrate the decomposition on accelerated magnetic resonance imaging (MRI), and finally apply the calibration diagnostics to electroencephalography (EEG) source imaging.
\end{abstract}

\section{Introduction}
Inverse problems are common in fields such as natural sciences, medical imaging, and astronomy. These problems are often ill-posed, in the sense that the measurement is many-to-one, where distinct underlying states can produce the same measurement. Consequently, reconstruction uncertainty is not only induced by measurement noise, but also arises from the geometry of the forward operator itself. In such settings, the observations identify only a subset of the underlying system, while the remaining degrees of freedom are fundamentally unresolvable from the data.

Classical optimization methods address ill-posed inverse problems by introducing explicit constraints, such as regularization that impose sparsity, and are typically evaluated by reconstruction metrics such as mean squared error (MSE). More recently, generative models aim to sample plausible reconstructions from a learned data prior, and are additionally evaluated by distributional metrics, such as Fréchet Inception Distance (FID)~\cite{heusel2017gans} 
and Contrastive Language--Image Pretraining (CLIP) score~\cite{radford2021learning}. However, they are limited because they assess either the reconstruction fidelity or the distributional similarity between generated samples and real data — without examining whether the model faithfully captures the posterior uncertainty over all the possible reconstructions consistent with a given measurement.
This raises a fundamental question: how should the uncertainty represented by these posterior samples be interpreted? In particular, existing uncertainty estimates typically conflate two qualitatively distinct sources of variability — ambiguity that is intrinsic to the ill-posed problem, and uncertainty that arises from measurement noise. Answering this question is especially consequential in high-stakes applications, where a model must not only produce accurate reconstructions but also faithfully reflect how much of its uncertainty is irreducible.

In this paper, we propose a decomposition of posterior uncertainty in inverse problems that separates intrinsic ambiguity from uncertainty propagated by noisy measurements. Building on this decomposition, we utilize a cascade generative framework that makes these components directly inspectable. This, in turn, provides diagnostic metrics that can complement reconstruction loss in model selection and parameter tuning. We validate the framework on a Gaussian example with known analytical posterior structure, accelerated MRI for qualitative visualization, and EEG source imaging as a realistic large-scale ill-posed inverse problem. Rather than performing an exhaustive benchmark, we use representative generative models to illustrate how uncertainty can be distributed differently between intrinsic ambiguity and noise-propagated variability, and how these differences lead to distinct qualitative and quantitative failure modes. In practice, this distinction helps determine whether uncertainty is dominated by irreducible non-identifiability or by factors that could be mitigated through better measurements, modeling, or inference. The factorized formulation also yields a modular perspective on posterior modeling: identifiable inference and ambiguity modeling. In this paper, we focus on the latter as the central object for uncertainty calibration.

\section{Background}
\textbf{Inverse Problems}
Inverse problems arise across domains, including medical imaging, astronomy, and image restoration~\cite{lustig2007sparse,starck2002deconvolution,dong2015image}. Classical methods address ill-posed inverse problems by imposing regularization or handcrafted constraints to select plausible reconstructions from an underdetermined solution set. More recently, generative methods have been used to learn distributions over possible reconstructions, including plug-and-play priors and diffusion posterior samplers~\cite{chung2022diffusion,daras2024survey}. We refer to models that directly sample full reconstructions $\hat{x}\sim p_\theta(x\mid y)$ as monolithic posterior models. Several diffusion inverse solvers further exploit the geometry of the forward operator. For example, DDRM~\cite{kawar2022denoising} uses spectral structure to construct restoration samplers, while DDNM~\cite{wang2022zero} uses a range--null decomposition to enforce measurement consistency and refine null-space content during reverse diffusion. Related null-space generative methods also learn conditional models for the unmeasured component in MRI~\cite{wen2023conditional}. These works use the geometry of the forward operator to construct posterior samples. We use the same geometric view for a different goal. The question is not only whether a model can generate plausible null-space components, but whether the induced conditional distribution over those components is calibrated. We therefore use the range--null view as a calibration tool rather than only as a restoration mechanism.

\textbf{Posterior uncertainty and calibration.} Recent work~\cite{qiu2026benchmarking} systematically evaluates posterior uncertainty in plug-and-play diffusion solvers and shows that solvers with comparable reconstruction quality can produce qualitatively different posterior variance. This raises a calibration question beyond whether posterior variability is large or small, but which component of uncertainty should be calibrated, and what distribution should it be calibrated against? A scalar posterior variance alone does not answer this question. Simulation-based calibration (SBC)~\cite{talts2018validating} instead tests whether posterior samples are statistically consistent with draws from the data-generating distribution. Across repeated simulations $x^*\sim p(x)$ and $y\sim p(y\mid x^*)$, a calibrated sampler should assign the true draw $x^*$ a uniform rank among samples from the approximate posterior.

\textbf{Our contribution}
We first ask what posterior uncertainty is composed of in a linear inverse problem. Our decomposition identifies intrinsic ambiguity as the principal target of calibration, reflecting uncertainty that is irreducible under the forward operator. We then use a cascade formulation to make this ambiguity directly accessible for both qualitative and quantitative calibration analysis.

\section{Decomposing Posterior Uncertainty}\label{sec:uncertainty_decomp}
We consider the standard linear inverse problem
\begin{equation}\label{eq:forward_model}
\mathbf{y} = \mathbf{Ax} + \boldsymbol{\varepsilon},
\end{equation} where $\mathbf{A} \in \mathbb{R}^{n \times p}$ denotes the forward operator, $\mathbf{x} \in \mathbb{R}^p$ is the unknown signal, and $\mathbf{y} \in \mathbb{R}^n$ is the measurement. Throughout the inverse problems considered here, the measurement dimension is typically much smaller than the signal dimension, i.e., $n \ll p$, which makes solving for $\mathbf{x}$ ill-posed. The term $\varepsilon$ denotes measurement noise, which is often modeled as Gaussian in practice.

In the linear setting, this decomposition is naturally induced by the geometry of the forward operator. Let
\(\mathbf{A} = \mathbf{U}\boldsymbol{\Sigma}\mathbf{V}^\top\) be the singular value decomposition (SVD) of \(\mathbf{A}\), with \(r=\mathrm{rank}(\mathbf{A})\), and partition the right singular vectors as \(\mathbf{V}=[\mathbf{V}_r,\mathbf{V}_n]\), where \(\mathbf{V}_r\in\mathbb{R}^{p\times r}\) and \(\mathbf{V}_n\in\mathbb{R}^{p\times(p-r)}\). The columns of \(\mathbf{V}_r\) span the identifiable subspace associated with nonzero singular values of \(\mathbf{A}\), and the columns of \(\mathbf{V}_n\) span \(\mathrm{Null}(\mathbf{A})\). Then any signal $\boldsymbol{x}$ can be written as
\begin{equation}\label{eq:recon_x}
\mathbf{x} = \mathbf{V}_r \boldsymbol{\alpha} 
+ \mathbf{V}_n \boldsymbol{\beta},
\end{equation}
where \(\boldsymbol{\alpha} \in \mathbb{R}^r\) and \(\boldsymbol{\beta} \in \mathbb{R}^{p-r}\) are coefficient vectors with respect to the bases \(\mathbf{V}_r\) and \(\mathbf{V}_n\), respectively. This decomposition is deterministic for any fixed \(\mathbf{x}\).
In the posterior inference problem, however, $\mathbf{x}$ is treated as a random variable with distribution $p(\mathbf{x}\mid\mathbf{y})$, and therefore 
the induced coefficients $\boldsymbol{\alpha}=\mathbf{V}_r^\top\mathbf{x}$ 
and $\boldsymbol{\beta}=\mathbf{V}_n^\top\mathbf{x}$ are random variables as well.
Substituting the decomposition into the forward model gives
\begin{equation}\label{eq:forward_decomp}
\mathbf{y} = \mathbf{A}(\mathbf{V}_r\boldsymbol{\alpha} + \mathbf{V}_n\boldsymbol{\beta}) + \boldsymbol{\varepsilon} = \mathbf{A}\mathbf{V}_r \boldsymbol{\alpha} + \boldsymbol{\varepsilon}.
\end{equation}
Thus, the measurement depends only on $\boldsymbol{\alpha}$, while $\boldsymbol{\beta}$ is invisible to the forward operator. In this sense, $\boldsymbol{\alpha}$ captures the measurement-identifiable component of $\mathbf{x}$, whereas $\boldsymbol{\beta}$ is intrinsically ambiguous.

The posterior inference problem can be equivalently expressed in terms of the coefficient vectors $(\boldsymbol{\alpha},\boldsymbol{\beta})$. Moreover, Eq.~\eqref{eq:forward_decomp} shows that the likelihood satisfies
\(p(\mathbf{y}\mid \boldsymbol{\alpha},\boldsymbol{\beta})
= p(\mathbf{y}\mid \boldsymbol{\alpha})\), assuming measurement noise independent of the signal. Therefore, the decomposed model implies the conditional independence relation $\boldsymbol{\beta} \perp \mathbf{y} \mid \boldsymbol{\alpha}$, under which the posterior factorizes as

\begin{equation}
p(\mathbf{x} \mid \mathbf{y})
= p(\boldsymbol{\alpha}, \boldsymbol{\beta} \mid \mathbf{y})
= p(\boldsymbol{\alpha} \mid \mathbf{y})\,
p(\boldsymbol{\beta} \mid \boldsymbol{\alpha}, \mathbf{y})
= p(\boldsymbol{\alpha} \mid \mathbf{y})\,
p(\boldsymbol{\beta} \mid \boldsymbol{\alpha}).
\label{eq:factorization}
\end{equation}

This factorization separates two qualitatively distinct forms of uncertainty. Uncertainty in $p(\boldsymbol{\alpha} \mid \mathbf{y})$ reflects noise and inference error, which can in principle be reduced by better measurements or stronger inference. By contrast, $\boldsymbol{\beta}$ is fundamentally unidentifiable from $\mathbf{y}$, and its uncertainty at fixed 
$\boldsymbol{\alpha}$ is the object of our calibration analysis. We define the intrinsic ambiguity at the oracle-identifiable state $\boldsymbol{\alpha}^*$ as:
\[
\mathcal{A}(\boldsymbol{\alpha}^*) := \mathbb{V}(\boldsymbol{\beta} \mid 
\boldsymbol{\alpha}^*),
\]
where $\mathbb{V}$ denotes the covariance matrix. This is the irreducible uncertainty induced by information loss in the forward operator. In practice, however, standard monolithic posterior samplers only expose $\mathbb{V}(\boldsymbol{\beta} \mid \mathbf{y})$. By the law of total variance and the conditional independence
$\boldsymbol{\beta} \perp \mathbf{y} \mid \boldsymbol{\alpha}$ from Eq.~\eqref{eq:factorization}, this quantity satisfies
\begin{align}
\mathbb{V}(\boldsymbol{\beta} \mid \mathbf{y})
&= \mathbb{E}_{p(\boldsymbol{\alpha} \mid \mathbf{y})}\!\left[
   \mathbb{V}(\boldsymbol{\beta} \mid \boldsymbol{\alpha}, \mathbf{y})\right]
+  \mathbb{V}_{p(\boldsymbol{\alpha} \mid \mathbf{y})}\!\left(
   \mathbb{E}[\boldsymbol{\beta} \mid \boldsymbol{\alpha}, \mathbf{y}]\right) 
   \notag \\
&= \underbrace{\mathbb{E}_{p(\boldsymbol{\alpha} \mid \mathbf{y})}\!\left[
   \mathbb{V}(\boldsymbol{\beta} \mid \boldsymbol{\alpha})\right]
   }_{\text{posterior-averaged intrinsic ambiguity}}
+  \underbrace{\mathbb{V}_{p(\boldsymbol{\alpha} \mid \mathbf{y})}\!\left(
   \mathbb{E}[\boldsymbol{\beta} \mid \boldsymbol{\alpha}]\right)
   }_{\text{propagated uncertainty from inferring } \boldsymbol{\alpha}}
\label{eq:total_var}
\end{align}
In the noiseless limit, $p(\boldsymbol{\alpha}\mid\mathbf{y})$ collapses to a point mass at $\boldsymbol{\alpha}^*$, the second term vanishes, and $\mathbb{V}(\boldsymbol{\beta}\mid\mathbf{y}) = \mathcal{A}(\boldsymbol{\alpha}^*)$. This indicates that in the idealized case with a noise-free measurement system that perfectly recovers the identifiable component, the remaining uncertainty is exactly the intrinsic ambiguity. This remaining uncertainty is not an inference error to be eliminated, but reflects the underdetermined nature of the inverse problem, and should therefore be calibrated rather than driven to zero. Hence, we identify the relevant calibration target as the conditional ambiguity distribution $p(\boldsymbol{\beta}\mid\boldsymbol{\alpha}^*)$, which monolithic posterior models do not directly expose. In the next section, we show how a cascade formulation makes this explicitly accessible.  


\section{Methods}
\subsection{Cascade Model}\label{sec:cascade}

Motivated by the factorization in Eq.~\eqref{eq:factorization}, we model the posterior in two stages: a range model for the measurement-identifiable component and a null model for the intrinsically ambiguous component. The resulting cascade architecture is shown in Figure~\ref{fig:cascade_posterior}, and the corresponding graphical model interpretation is shown in Appendix~\ref{app:pgm}.

\textbf{Range model}
The range model approximates the conditional object $p(\boldsymbol{\alpha} \mid \mathbf{y})$ and is responsible for inference of the measurement-identifiable component. In the supervised training setting considered here, oracle $\boldsymbol{\alpha}^*$ is available because the ground-truth $\mathbf{x}$ is known and can be projected onto the identifiable subspace. During training, $\boldsymbol{\alpha}^*$ serves as the supervision target for the range model. At inference time, only $\mathbf{y}$ is observed, and the model produces an estimate or distribution over $\boldsymbol{\alpha}$. This stage may be implemented either probabilistically or deterministically, with the deterministic case viewed as a point-mass approximation at $\hat{\boldsymbol{\alpha}}(\mathbf{y})$.

\textbf{Null model}
The null model approximates $p(\boldsymbol{\beta} \mid \boldsymbol{\alpha})$ and is responsible for the conditional generation of the intrinsically ambiguous component. During training, oracle $\boldsymbol{\alpha}^*$ is used as the conditioning input and the corresponding $\boldsymbol{\beta}^*$ as the target output. This gives direct access to the model conditional $p^\mathrm{null}_{\boldsymbol{\phi}}(\boldsymbol{\beta} \mid \boldsymbol{\alpha}^*)$, which is the key object for calibration analysis of intrinsic ambiguity. Expressive generative modeling is most critical in this stage, since the intrinsic ambiguity that remains after fixing $\boldsymbol{\alpha}^*$ must be represented here.

\textbf{Inference and Evaluation}
At inference time, posterior samples are generated in cascade form by first obtaining $\hat{\alpha}$ from the range model and then sampling $\hat{\boldsymbol{\beta}} \sim p^\mathrm{null}_\phi(\boldsymbol{\beta} \mid \hat{\boldsymbol{\alpha}})$. The final prediction $\hat{\mathbf{x}}$ is reconstructed according to Eq.~\eqref{eq:recon_x}. During evaluation, oracle $\boldsymbol{\alpha}^*$ from the test set is used only for calibration diagnostics, whereas reconstruction quality is assessed on end-to-end cascade predictions formed from inferred $\hat{\boldsymbol{\alpha}}$ and sampled $\hat{\boldsymbol{\beta}}$.
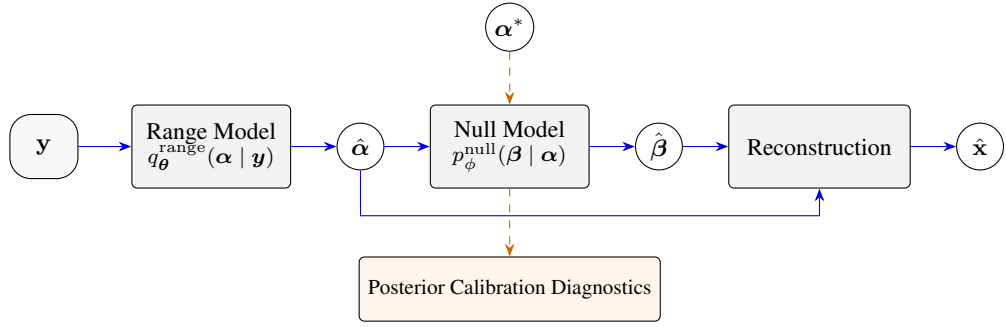
\begin{figure*}[t]
\centering
\begin{tikzpicture}[
font=\small,
>=Stealth,
node distance=0.7cm and 0.7cm,
box/.style={draw, rounded corners=2pt, thin, fill=black!5, minimum height=1.1cm, minimum width=2.1cm, align=center},
diagbox/.style={draw, rounded corners=2pt, thin, fill=orange!8, minimum height=0.85cm, minimum width=2.6cm, align=center, font=\footnotesize},
io/.style={draw, rounded corners=8pt, thin, fill=black!3, minimum height=0.85cm, minimum width=0.9cm, align=center},
latent/.style={draw, thin, circle, inner sep=1.8pt, minimum size=0.62cm, fill=white},
inf/.style={->, thin, draw=blue!85!black},
oracle/.style={->, thin, dashed, draw=orange!80!black}
]
\node[io] (y) {$\mathbf{y}$};
\node[box, right=0.7cm of y] (range) {Range Model\\[-1pt]\footnotesize $q_{\boldsymbol{\theta}}^\mathrm{range}(\boldsymbol{\alpha} \mid \boldsymbol{y})$};
\node[latent, right=0.6cm of range] (ahat) {$\hat{\boldsymbol{\alpha}}$};
\node[box, right=0.6cm of ahat] (null) {Null Model\\[-1pt]\footnotesize $p_{\phi}^\mathrm{null}(\boldsymbol{\beta} \mid \boldsymbol{\alpha})$};
\node[latent, right=0.6cm of null] (beta) {$\hat{\boldsymbol{\beta}}$};
\node[box, right=0.6cm of beta, minimum width=2.4cm] (recon) {Reconstruction};
\node[latent, right=0.6cm of recon] (x) {$\hat{\mathbf{x}}$};

\node[latent, above=0.7cm of null] (astar) {$\boldsymbol{\alpha}^*$};

\node[diagbox, below=0.9cm of null] (diag) 
{Posterior Calibration Diagnostics};

\draw[inf] (y) -- (range);
\draw[inf] (range) -- (ahat);
\draw[inf] (ahat) -- (null);
\draw[inf] (null) -- (beta);
\draw[inf] (beta) -- (recon);
\draw[inf] (recon) -- (x);

\draw[inf] (ahat.south) -- ++(0,-0.6) -| (recon.south);

\draw[oracle] (astar) -- (null);
\draw[oracle] (null) -- (diag);

\end{tikzpicture}
\caption{\textbf{Cascade architecture.} The null model sits at the intersection of two operational paths. \textcolor{blue!85!black}{Inference (blue, horizontal):} given $\mathbf{y}$, the range model produces $\hat{\boldsymbol{\alpha}}$ and the null model produces $\hat{\boldsymbol{\beta}} \mid \hat{\boldsymbol{\alpha}}$; reconstruction combines both to yield $\hat{\mathbf{x}}$. \textcolor{orange!80!black}{Diagnostic (orange dashed, vertical):} the null model is queried at oracle $\boldsymbol{\alpha}^*$ to expose $p^\mathrm{null}_\phi(\boldsymbol{\beta} \mid \boldsymbol{\alpha}^*)$ for posterior calibration. At training time, both modules use oracle $\boldsymbol{\alpha}^*$: the range model on $(\mathbf{y}, \boldsymbol{\alpha}^*)$ pairs, the null model on $(\boldsymbol{\alpha}^*, \boldsymbol{\beta}^*)$ pairs.}
\label{fig:cascade_posterior}
\end{figure*}


\subsection{Posterior Calibration Metric}\label{sec:SBC_intro}
\textbf{Simulation-based calibration.}
The decomposition in Section~\ref{sec:uncertainty_decomp} identifies $p^\mathrm{null}_\phi(\boldsymbol{\beta} \mid \boldsymbol{\alpha}^*)$ as the relevant object for assessing whether a model is well calibrated in terms of intrinsic ambiguity. Here oracle $\boldsymbol{\alpha}^*$ is obtained from the test set and is used only for calibration diagnostics. Ideally, one would like to directly compare the full model distribution $p^\mathrm{null}_\phi(\boldsymbol{\beta} \mid \boldsymbol{\alpha}^*)$ against the corresponding data-distribution $p(\boldsymbol{\beta} \mid \boldsymbol{\alpha}^*)$. In practice, however, direct calibration in this high-dimensional space is intractable. We therefore use simulation-based calibration (SBC), a diagnostic for posterior calibration based on repeated simulation from the generative model~\cite{talts2018validating}, recently adopted for amortized diffusion posteriors in simulation-based inference~\cite{nautiyal2025condisim}. Under calibration, a true draw $\boldsymbol{\beta}^* \sim p(\boldsymbol{\beta} \mid \boldsymbol{\alpha}^*)$ should be statistically indistinguishable from samples $\tilde{\boldsymbol{\beta}}^{(1)},\dots,\tilde{\boldsymbol{\beta}}^{(L)} \sim p^\mathrm{null}_\phi(\boldsymbol{\beta} \mid \boldsymbol{\alpha}^*)$. Accordingly, in SBC, the rank of a chosen scalar summary $T(\boldsymbol{\beta}^*)$ among $\{T(\tilde{\boldsymbol{\beta}}^{(\ell)})\}_{\ell=1}^L$ should be uniformly distributed. Deviations from uniformity indicate miscalibration. This provides a necessary but not sufficient condition for correct posterior calibration.

Since SBC requires scalar test statistics, we evaluate calibration through interpretable functionals of the ambiguous component $\boldsymbol{\beta}$, rather than attempting to diagnose calibration separately in every coordinate. We introduce the Euclidean norm $\|\boldsymbol{\beta}\|_2$ and the peak-to-total ratio $\|\boldsymbol{\beta}\|_\infty / \|\boldsymbol{\beta}\|_2$ as our test statistics (see Appendix~\ref{app:sbc-high-dim} for a discussion on the challenges of per-dimensional SBC in high dimensions). In particular, the statistic $\|\boldsymbol{\beta}\|_2$ captures the overall magnitude of the measurement-invisible component, while $\|\boldsymbol{\beta}\|_\infty / \|\boldsymbol{\beta}\|_2$ characterizes how that magnitude is distributed across dimensions. We use $\|\boldsymbol{\beta}\|_2$ rather than $\|\boldsymbol{\beta}\|_2^2$ simply for consistency with the second statistic. For SBC, both choices are equivalent since they yield identical ranks. The squared norm nevertheless provides a useful interpretation:
$\mathbb{E}\!\left[\|\boldsymbol{\beta}\|_2^2 \mid \boldsymbol{\alpha}^*\right]
=
\left\|\mathbb{E}\!\left[\boldsymbol{\beta} \mid \boldsymbol{}\alpha^*\right]\right\|_2^2
+
\operatorname{tr}\!\left(\mathbb{V}(\boldsymbol{\beta} \mid \boldsymbol{\alpha}^*)\right)$. Thus, when $\mathbb{E}[\boldsymbol{\beta} \mid \boldsymbol{\alpha}^*]=0$, $\|\boldsymbol{\beta}\|_2^2$ reduces in expectation to the total conditional variance across dimensions. Together, these two statistics probe complementary aspects of whether $p^\mathrm{null}_\phi(\boldsymbol{\beta} \mid \boldsymbol{\alpha}^*)$ is calibrated to the corresponding data-generating distribution $p(\boldsymbol{\beta} \mid \boldsymbol{\alpha}^*)$.

\textbf{Qualitative diagnostic.}
SBC provides a quantitative test of calibration but requires repeated simulation across many values of $\boldsymbol{\alpha}^*$. As a complementary qualitative diagnostic, the cascade also exposes an intrinsic ambiguity map in the original signal space. Given an oracle identifiable component $\boldsymbol{\alpha}^*$, we draw $K$ samples $\boldsymbol{\beta}^{(k)} \sim p^\mathrm{null}_\phi(\boldsymbol{\beta} \mid \boldsymbol{\alpha}^*)$, reconstruct them in the original coordinates as $\mathbf{x}^{(k)} = \mathbf{V_r} \boldsymbol{\alpha}^* + \mathbf{V_n} \boldsymbol{\beta}^{(k)}$, and compute the per-coordinate empirical variance
\begin{equation}
\widehat{\mathbb{V}}_k[\mathbf{x}^{(k)}]
\;\xrightarrow{K \to \infty}\;
\mathrm{diag}\!\left(\mathbf{V_n} \,\mathbb{V}
(\boldsymbol{\beta} \mid \boldsymbol{\alpha}^*)\, \mathbf{V_n}^\top\right).
\end{equation}
We refer to this quantity as the intrinsic ambiguity map. In settings where the original-space coordinate system is shared across test cases (e.g., a fixed cortical mesh in EEG), these maps may also be averaged across $\boldsymbol{\alpha}^* \sim p_{\mathrm{test}}$ to obtain a population-level view. When coordinates differ across examples (e.g., MRI slices or anatomies), the map is inspected on a per-example basis.

A calibrated null model should place intrinsic ambiguity in geometrically plausible regions of the signal space, providing a useful visual sanity check for domain experts. For example, in accelerated MRI one expects higher ambiguity in structures poorly constrained by the retained Fourier coefficients, while in EEG one expects larger ambiguity in deep or midline brain sources. This diagnostic is complementary to SBC: the spatial map reflects \emph{marginal} per-coordinate variance, whereas SBC probes scalar summary statistics of the \emph{joint} conditional distribution $p^\mathrm{null}_\phi(\boldsymbol{\beta} \mid \boldsymbol{\alpha}^*)$, such as total energy or concentration.

\section{Experiments}
\subsection{Gaussian Analytical Example}
We first study a controlled Gaussian example in which the relevant conditional distributions are analytically tractable. This provides a sanity check for both the proposed decomposition and the calibration diagnostics. Unlike in realistic inverse problems, the oracle ambiguity distribution $p(\boldsymbol{\beta} \mid \boldsymbol{\alpha}^*)$ is available in closed form in this setting, allowing direct comparison between the learned null model and the true conditional target.
\begin{figure}
\centering
\includegraphics[width=0.85\linewidth]{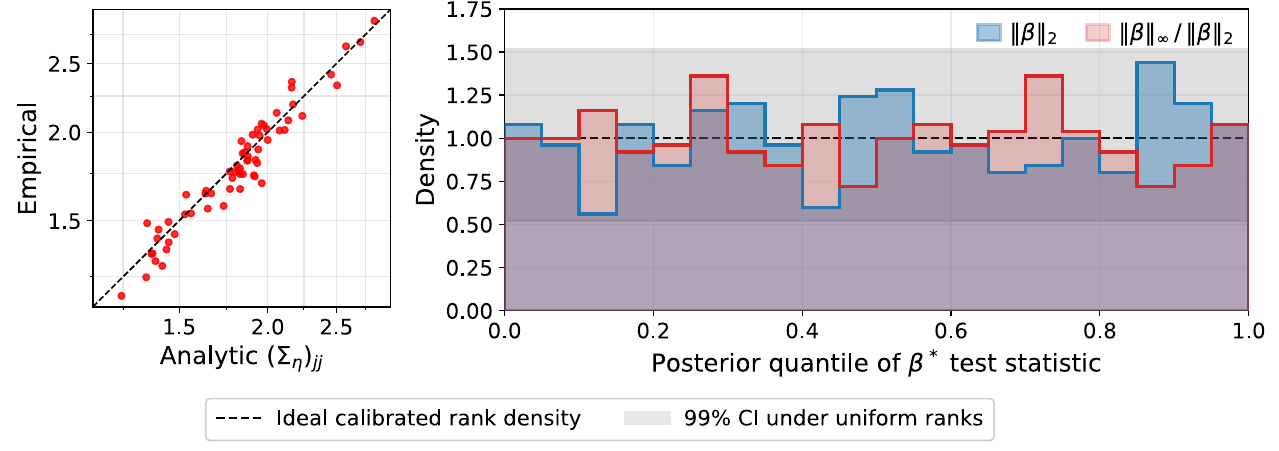}
\caption{\textbf{Left:} Per-dimension $\boldsymbol{\beta}$ variance calibration. Each dot corresponds to one of the $q=64$ $\boldsymbol{\beta}$ dimensions and compares the empirical variance of DDPM samples from $p^\mathrm{null}_{\phi}(\boldsymbol{\beta} \mid \boldsymbol{\alpha}^*)$ with the corresponding analytical variance from $\mathrm{diag}(\boldsymbol{\Sigma}_{\boldsymbol{\eta}})$. The dashed line indicates equality. The mean empirical-to-analytical variance ratio across dimensions is $0.985$, indicating reasonable calibration. \textbf{Right:} SBC diagnostics. Rank histograms for the test statistics $\|\boldsymbol{\beta}\|_2$ and $\|\boldsymbol{\beta}\|_\infty / \|\boldsymbol{\beta}\|_2$ under the learned DDPM null model. Ranks are normalized to $[0,1]$ and plotted as densities, so the ideal calibrated density is $1$. The gray band shows the per-bin $99\%$ binomial confidence interval under uniform ranks. Histograms are computed with $200$ posterior samples for each of the $500$ test cases.}
\label{fig:gaussian_calib}
\end{figure}

We generate data as a simple linear Gaussian instantiation of the factorization in Eq.~\eqref{eq:factorization}, where $\boldsymbol{y}$ depends only on $\boldsymbol{\alpha}$ by construction, so that $\boldsymbol{\beta} \perp \boldsymbol{y} \mid \boldsymbol{\alpha}$ holds exactly (see Appendix~\ref{app:gaussian_details} for parameter construction and 
training details):
\begin{align}
\boldsymbol{\alpha} &\sim \mathcal{N}(0, \mathbf{I}_r), \\
\boldsymbol{\beta} \mid \boldsymbol{\alpha} &\sim 
\mathcal{N}(\mathbf{C}\boldsymbol{\alpha}, \boldsymbol{\Sigma}_{\boldsymbol{\eta}}), \\
\boldsymbol{y} \mid \boldsymbol{\alpha} &\sim 
\mathcal{N}(\mathbf{A}\boldsymbol{\alpha}, \sigma_y^2 \mathbf{I}_n),
\end{align}
where $\boldsymbol{\alpha} \in \mathbb{R}^r$ is the identifiable component, $\boldsymbol{\beta} \in \mathbb{R}^q$ is the ambiguous component, and $\mathbf{y} \in \mathbb{R}^n$ is the measurement. Here $\mathbf{A} \in \mathbb{R}^{n \times r}$ is the forward operator on the identifiable component, $\mathbf{C} \in \mathbb{R}^{q \times r}$ controls the dependence of the ambiguous component on $\boldsymbol{\alpha}$, and $\boldsymbol{\Sigma}_{\boldsymbol{\eta}} \in \mathbb{R}^{q \times q}$ specifies the intrinsic ambiguity covariance. By construction, the measurement depends only on $\boldsymbol{\alpha}$, while $\boldsymbol{\beta}$ remains unobserved. In our experiments, we use $r=32$, $q=64$, and $n=32$. The oracle ambiguity distribution is therefore $\boldsymbol{\beta} \mid \boldsymbol{\alpha}^* \sim \mathcal{N}(\mathbf{C}\boldsymbol{\alpha}^*, \boldsymbol{\Sigma}_{\boldsymbol{\eta}})$, with conditional covariance $\mathbb{V}(\boldsymbol{\beta} \mid \boldsymbol{\alpha}^*) = \boldsymbol{\Sigma}_{\boldsymbol{\eta}}$.

To ensure a nontrivial calibration target, $\boldsymbol{\Sigma}_{\boldsymbol{\eta}}$ 
is chosen with full off-diagonal structure, requiring the null model to match not only marginal variances but also cross-dimensional correlations. We then train a conditional DDPM~\cite{ho2020denoising} null model as a representative expressive sampler for $p^\mathrm{null}_\phi(\boldsymbol{\beta} \mid \boldsymbol{\alpha})$ and evaluate it at oracle $\boldsymbol{\alpha}^*$. Figure~\ref{fig:gaussian_calib} shows close agreement between the empirical per-dimension variance from DDPM samples and the analytical conditional variance, while the SBC histograms remain close to uniform for the test statistics $\|\boldsymbol{\beta}\|_2$ and $\|\boldsymbol{\beta}\|_\infty / \|\boldsymbol{\beta}\|_2$. This controlled example therefore serves as a sanity check: in a setting where the conditional ambiguity distribution is known analytically, a DDPM null model is reasonably calibrated against the exact target, and the proposed diagnostics correctly reflect this behavior.

\subsection{Accelerated MRI}
We use accelerated MRI as a visually intuitive setting to illustrate the posterior variance decomposition introduced in Section~\ref{sec:uncertainty_decomp}. In this setting, a standard monolithic posterior model would expose only the total posterior uncertainty, which conflates intrinsic ambiguity with uncertainty propagated from noisy inference. Because the identifiable component is available in closed form under Fourier undersampling, accelerated MRI allows these two contributions to be separated explicitly and visualized in image space.

The forward operator is $\mathbf{A}=\mathbf{M}\mathbf{F}$, where $\mathbf{F} \in \mathbb{C}^{N_v \times N_v}$ is the two-dimensional discrete Fourier transform and $\mathbf{M} \in \mathbb{R}^{N_v \times N_v}$ is a diagonal binary mask retaining a subset of Fourier coefficients. Measurements are \(\mathbf{y}_\sigma = \mathbf{M}\mathbf{F}\mathbf{x} + \boldsymbol{\varepsilon}_\sigma\), where \(\boldsymbol{\varepsilon}_\sigma \sim \mathcal{N}(0,\sigma^2 \mathbf{I})\) denotes measurement noise in k-space. For a clean image $\mathbf{x}$, we define the oracle identifiable component as $\mathbf{x}_\alpha^* = \mathbf{F}^{-1}\mathbf{M}\mathbf{F}\mathbf{x}$ and the null component as $\mathbf{x}_\beta = \mathbf{x} - \mathbf{x}_\alpha^*$. This is the identifiable/null decomposition from Section~\ref{sec:uncertainty_decomp}, expressed directly in image space rather than in singular-vector coordinates. By construction, $\mathbf{A}\mathbf{x}_{\boldsymbol{\beta}} = \mathbf{M}\mathbf{F}(\mathbf{x}-\mathbf{F}^{-1}\mathbf{M}\mathbf{F}\mathbf{x})=\mathbf{0}$, so $\mathbf{x}_{\boldsymbol{\beta}}$ lies in the null space of $\mathbf{A}$. In the noise-free case, $\mathbf{x}_\alpha^* = \mathbf{F}^{-1}\mathbf{y}$. Under noisy measurements, however, the available closed-form estimate is $\hat{\mathbf{x}}_\alpha = \mathbf{F}^{-1}\mathbf{y}_\sigma = \mathbf{x}_{\mathbf{\alpha}^*} + \mathbf{F}^{-1} \boldsymbol{\varepsilon}_\sigma$.

\begin{figure*}
\centering
\includegraphics[width=\linewidth]{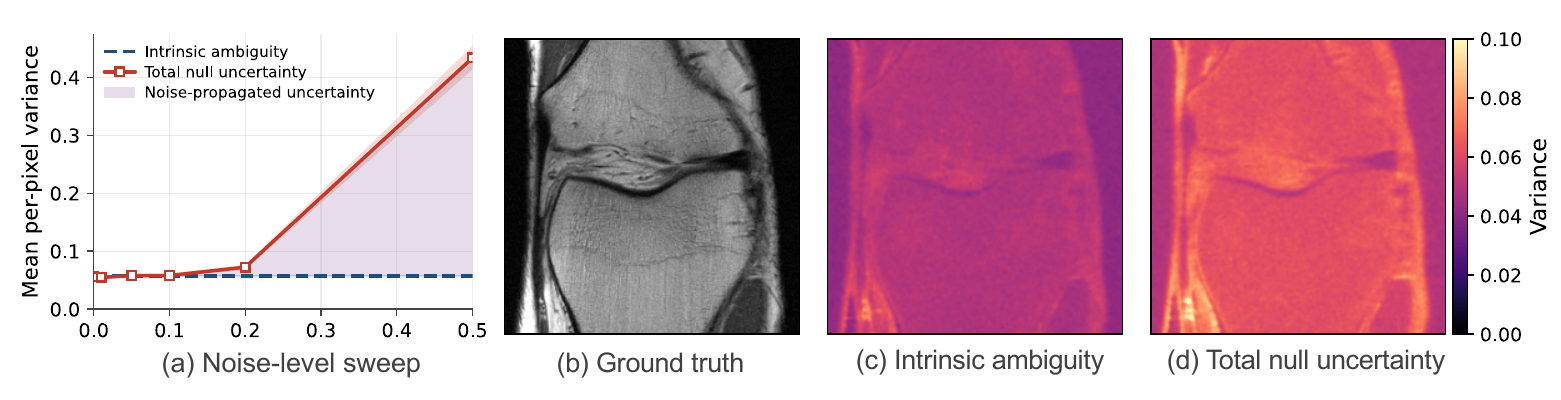}
\caption{Posterior uncertainty decomposition for accelerated MRI. Panel (a) shows a noise-level sweep over 50 held-out slices, with each point estimated by Monte Carlo sampling. Intrinsic ambiguity remains constant, while total null uncertainty increases with k-space noise. The shaded gap indicates noise-propagated uncertainty. Panel (b) shows a representative ground-truth slice. Panel (c) shows intrinsic ambiguity, estimated as $\mathbb{V}_{p_\phi}(\mathbf{x}_\beta \mid \mathbf{x}_\alpha^*)$ by fixing the oracle identifiable component. Panel (d) shows total null uncertainty at $\sigma=0.2$, estimated by marginalizing over noisy identifiable estimates $\hat{x}_\alpha = F^{-1}y_\sigma$. Both maps share the same color scale and report per-pixel variance in image intensity.}
\label{fig:MRI_decomp}
\end{figure*}

Because $\mathbf{x}_{\boldsymbol{\alpha}}^*$ is available in closed form for clean training images, we do not train a separate range model and only train a conditional diffusion model $p_\phi^\mathrm{null}(\mathbf{x}_{\boldsymbol{\beta}} \mid 
\mathbf{x}_{\boldsymbol{\alpha}})$ using oracle training pairs $(\mathbf{x}_{\boldsymbol{\alpha}}^*, \mathbf{x}_{\boldsymbol{\beta}}^*)$. This gives a clean setting in which propagated uncertainty is controlled directly by measurement noise, rather than by details of range-model training. For a unitary Fourier transform $\mathbf{F}$, and under a Lipschitz assumption on the conditional null mean (i.e. $m(\mathbf{x}_{\boldsymbol{\alpha}}) 
= \mathbb{E}_{p_\phi}[\mathbf{x}_{\boldsymbol{\beta}} 
\mid \mathbf{x}_{\boldsymbol{\alpha}}]$), the propagated uncertainty term in Eq.~\eqref{eq:total_var} is bounded by $O(\sigma^2)$; see Appendix~\ref{app:mri_bound}.

We use single-coil knee images from the NYU fastMRI dataset~\cite{knoll2020fastmri,zbontar2018fastmri} with an undersampling mask retaining 25\% of Fourier coefficients. Figure~\ref{fig:MRI_decomp} visualizes the model-predicted uncertainty decomposition. The noise-level sweep in Fig.~\ref{fig:MRI_decomp}(a) summarizes this decomposition across held-out slices by averaging the per-pixel variance over all pixels and slices. Intrinsic ambiguity remains approximately flat as the k-space noise level increases. This is expected because the intrinsic ambiguity map defined in Section~\ref{sec:SBC_intro} conditions on the oracle identifiable component $\mathbf{x}_{\boldsymbol{\alpha}}^*$, and is therefore independent of measurement noise. It reflects ambiguity induced by the forward operator and the data distribution. In contrast, total null uncertainty increases as measurement noise corrupts the identifiable estimate $\hat{\mathbf{x}}_\alpha = \mathbf{F}^{-1}\mathbf{y}$ and propagates into the conditional null distribution. The shaded region visualizes this noise-propagated uncertainty.

Figures~\ref{fig:MRI_decomp}(b)--(d) show this decomposition spatially for a representative slice. Panel (b) shows the ground-truth image. Panel (c) shows the model-predicted intrinsic ambiguity map described in Section~\ref{sec:SBC_intro}, corresponding to $\mathbb{V}_{p_\phi}(\mathbf{x}_{\boldsymbol{\beta}} \mid \mathbf{x}_{\boldsymbol{\alpha}}^*)$. Panel (d) shows the model-predicted total null uncertainty, corresponding to $\mathbb{V}_{p_\phi}(\mathbf{x}_{\boldsymbol{\beta}} \mid \mathbf{y})$ after marginalizing over noisy identifiable estimates $\hat{\mathbf{x}}_{\boldsymbol{\alpha}}$. If only panel (d) were available, as in a monolithic posterior sampler, the source of uncertainty would be unclear. The decomposition gives this uncertainty a physical interpretation. Panel (c) reflects uncertainty due to information loss from the forward operator and is therefore tied to the k-space undersampling protocol. The additional uncertainty in panel (d) reflects noise propagated through $\hat{\mathbf{x}}_{\boldsymbol{\alpha}}$, which can in principle be reduced by denoising, repeated acquisition, or better measurement.

\subsection{EEG Source Imaging}\label{sec:eeg}
We next evaluate our calibration analysis and cascade framework on EEG source imaging, a severely ill-posed inverse problem that provides a stress test for posterior uncertainty. The task is to recover high-resolution voxel-wise brain activity from sparse scalp measurements~\cite{michel2019eeg}. Since many distinct source configurations can explain the same EEG measurement, classical methods rely heavily on regularization, thereby imposing strong inductive biases on the reconstructions. We consider the standard linear forward model, $\mathbf{y} = \mathbf{Ax} + \boldsymbol{\varepsilon}$, where $\mathbf{x} \in \mathbb{R}^{N_v}$ denotes voxel-wise source activity, $\mathbf{y} \in \mathbb{R}^{N_s}$ denotes scalp measurements, and $\mathbf{A} \in \mathbb{R}^{N_s \times N_v}$ is the leadfield matrix. Because $N_s \ll N_v$ in practice, the forward operator has a large null space, and the decomposition from Section~\ref{sec:uncertainty_decomp} is especially relevant in EEG source imaging: only a low-dimensional component of source activity is measurement-identifiable, while a large component remains intrinsically ambiguous.

We construct a single-subject head model from the MNE sample dataset~\cite{gramfort2013meg}, based on a realistic MRI-derived anatomy. The leadfield matrix is computed via a three-layer boundary element method (BEM) on a cortical source space with fixed-orientation dipoles along the cortical normal, yielding $\mathbf{A} \in \mathbb{R}^{60 \times 4699}$ (i.e.\ 60 EEG sensors and 4699 cortical sources). The linear forward model is exact in the quasi-static regime of Maxwell's equations~\cite{hamalainen1993magnetoencephalography}. Source activity is simulated as spatially localized Gaussian patches with random central vertices, widths $\sigma \sim \mathrm{Uniform}(5, 20)$ mm, amplitudes $\sim \mathrm{Uniform}(0.5, 2.0)$, and signs flipped with probability $0.5$. The number of patches is sampled uniformly from $\{1,2,3\}$ for each example. Measurements are generated with added Gaussian noise calibrated to signal-to-noise ratio (SNR) $= 5$.

\begin{figure*}
\centering
\includegraphics[width=0.85\linewidth]{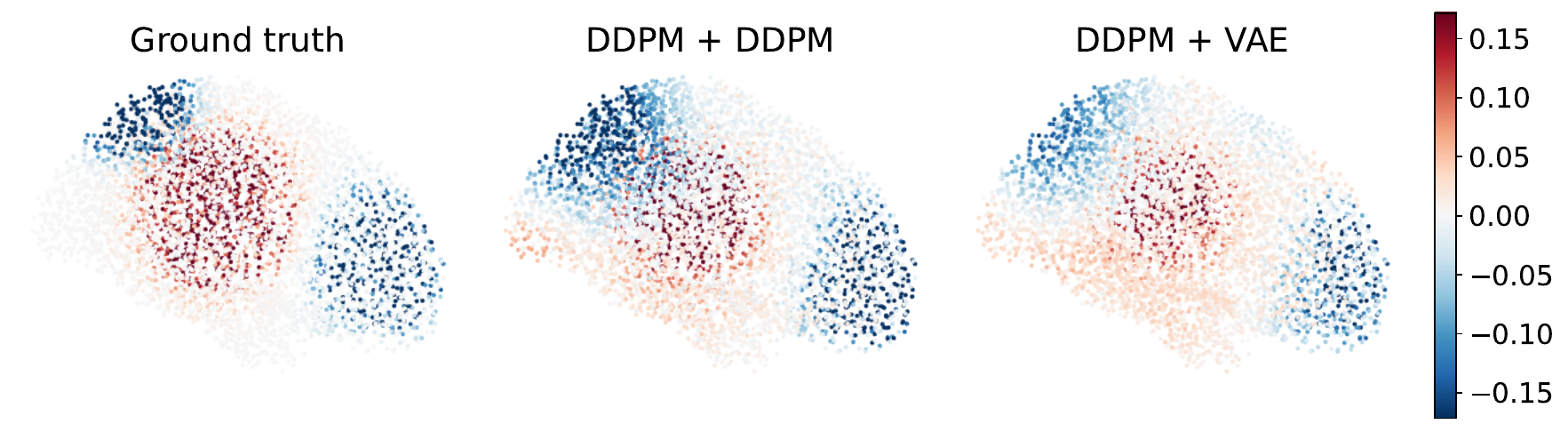}
\caption{Example source reconstruction for a held-out three-patch test case. Both cascades produce visually similar posterior mean reconstructions and recover the dominant spatial pattern of the ground truth, motivating our comparison of their intrinsic-ambiguity calibration beyond standard reconstruction quality. Color indicates signed source amplitude.}
\label{fig:EEG_recon}
\end{figure*}

Following the cascade strategy, we independently train multiple range and null models, then evaluate their combinations, as described in Section~\ref{sec:cascade}. The range model is instantiated as a DDPM, variational autoencoder (VAE), or multilayer perceptron (MLP), while the null model is instantiated as either a DDPM or VAE. We evaluate all range--null combinations on $450$ held-out test cases. The main text focuses on two cascades that share the same DDPM range model but differ in the null model: \emph{cascade A} pairs the DDPM range model with a DDPM null model, whereas \emph{cascade B} pairs the DDPM range model with a VAE null model at Kullback--Leibler (KL) weight $\lambda_\mathrm{KL}=1.0$~\cite{Higgins2016betaVAELB}. We use the posterior sample mean over $200$ samples as the point reconstruction, which is the standard Bayes estimator under squared-error loss and reflects the average behavior of the trained model. Reconstruction quality is measured by Pearson correlation between the posterior mean reconstruction and the ground-truth source signal, a measurement of spatial pattern similarity commonly used in EEG source-imaging evaluations~\cite{sun2023deep}. The two cascades achieve reasonable reconstructions even under the challenging three-patch source configuration, as shown in Fig.~\ref{fig:EEG_recon}, and have nearly identical Pearson correlation with the ground truth ($0.698$ vs.\ $0.693$). We therefore ask whether models that are nearly indistinguishable under standard reconstruction metrics differ in their calibration of intrinsic ambiguity. Remaining range--null combinations and architecture details are deferred to Appendix~\ref{sec:additional_res}.

\begin{figure*}
\centering
\includegraphics[width=0.85\linewidth]{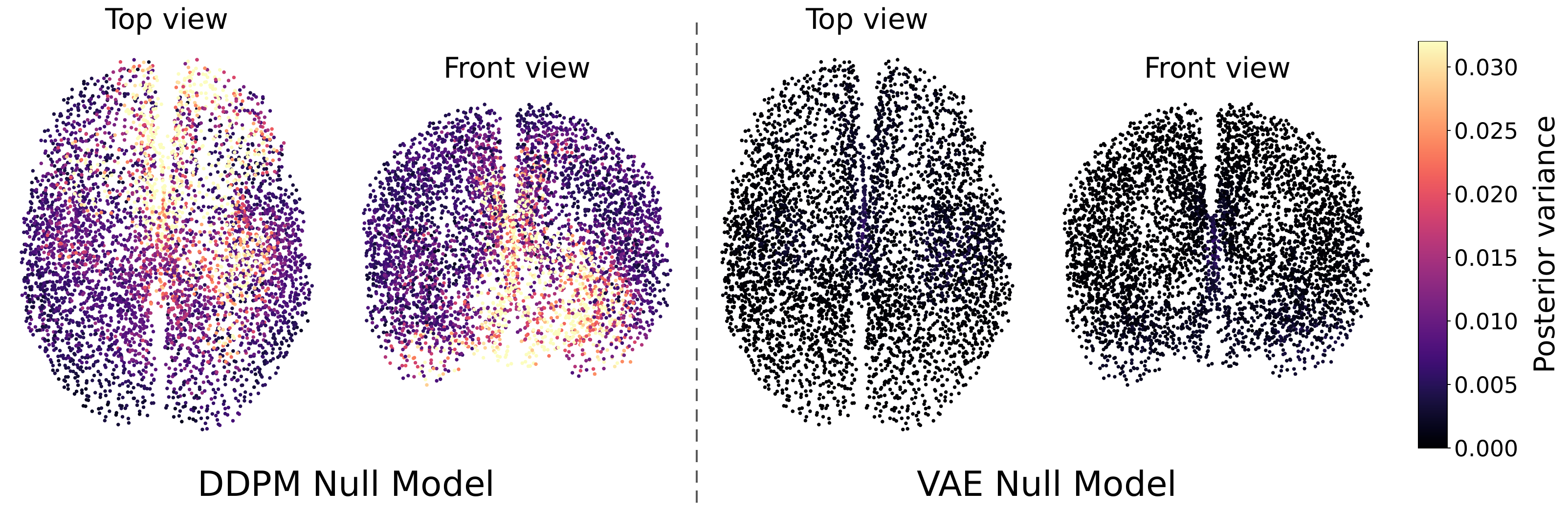}
\caption{\textbf{Averaged intrinsic ambiguity map on the cortical surface.} Per-voxel intrinsic ambiguity map averaged across $450$ test cases ($M=200$ posterior samples per case). \textbf{Left:} DDPM null model produces variance concentrated along the cortical midline and at deep medial regions. \textbf{Right:} VAE null model produces variance that is uniformly low across the cortex. Both models share a linear color scale capped at the 95th percentile of the DDPM map.}
\label{fig:EEG_cortex}
\end{figure*}

\textbf{Qualitative diagnostic.} We plot the averaged intrinsic ambiguity map (calculated according to Section~\ref{sec:SBC_intro}) on the cortical surface as a qualitative calibration check, contrasting the DDPM and VAE null models. As shown in Figure~\ref{fig:EEG_cortex}, the DDPM null produces larger intrinsic ambiguity at deeper brain voxels and along the longitudinal fissure, qualitatively consistent with the well-known reduced sensitivity of EEG to deep and medial sources~\cite{hauk2022towards,chamanzar2021neural}. The VAE null, by contrast, produces a uniformly low-variance profile across all brain regions, indicating that the model assigns negligible intrinsic ambiguity regardless of source location. This visual under-dispersion already signals miscalibration of $p^\mathrm{null}_\phi(\boldsymbol{\beta} \mid \boldsymbol{\alpha}^*)$, motivating a quantitative test.

\begin{figure*}
\centering
\includegraphics[width=0.8\linewidth]{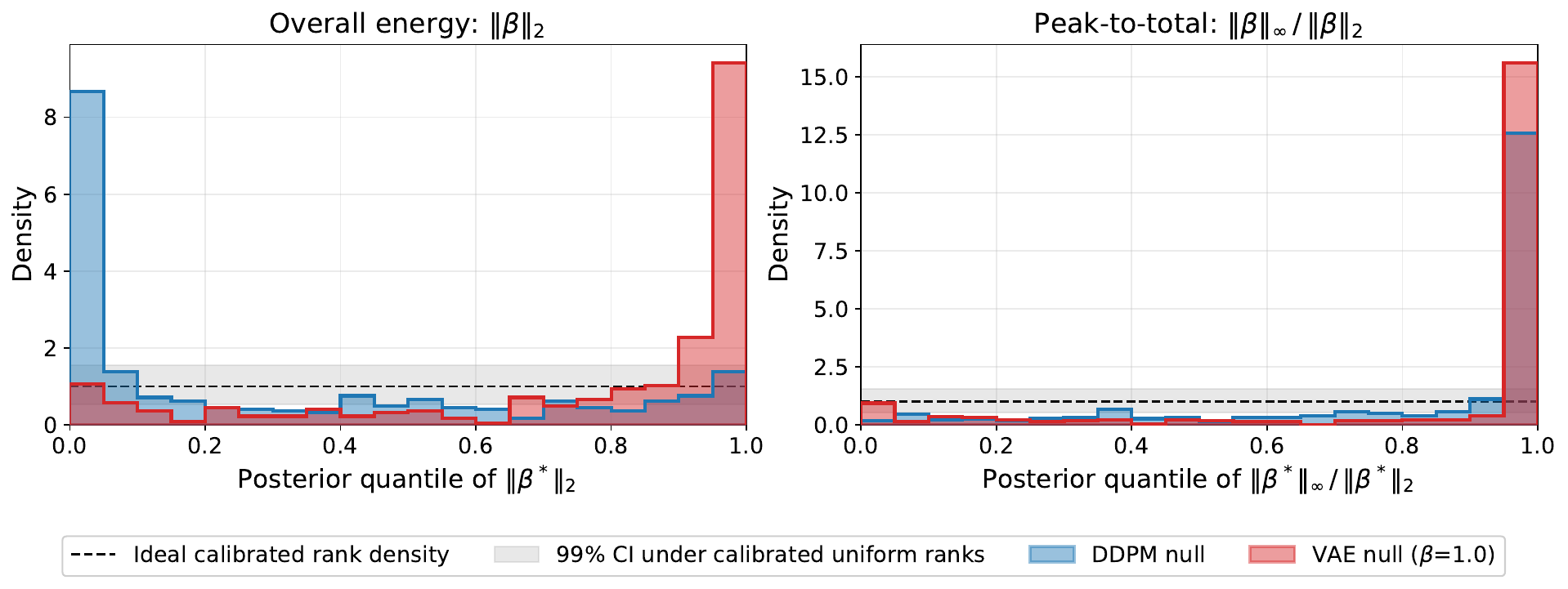}
\caption{\textbf{Similar reconstruction does not imply similar calibration of intrinsic ambiguity.}
With the DDPM range model fixed, the DDPM null and VAE null models achieve similar posterior-mean reconstruction quality, yet their null-space SBC histograms differ substantially. Left: posterior quantile of the true null-space energy $\|\beta^*\|_2$. Right: posterior quantile of the true peak-to-total ratio $\|\beta^*\|_\infty / \|\beta^*\|_2$. The DDPM null has higher null-space energy, while the VAE null is under-energetic.  Both models are more isotropic than the true data distribution.}
\label{fig:EEG_SBC}
\end{figure*}
\textbf{SBC Quantitative diagnostic.} Figure~\ref{fig:EEG_SBC} reports SBC histograms at oracle $\boldsymbol{\alpha}^*$ for the DDPM and VAE null models, using the geometric test statistics $\|\boldsymbol{\beta}\|_2$ and $\|\boldsymbol{\beta}\|_\infty / \|\boldsymbol{\beta}\|_2$ introduced in Section~\ref{sec:SBC_intro}. Neither null model yields uniform histograms; the deviations exceed the $99\%$ confidence interval expected under perfect calibration, revealing two qualitatively distinct failure modes. On total energy $\|\boldsymbol{\beta}\|_2$, the two models fail in opposite directions: the DDPM null concentrates rank mass near zero, indicating that its samples systematically carry more energy than the true data-generating distribution, while the VAE null concentrates rank mass near one, indicating collapse toward values smaller than the truth. On peak-to-total $\|\boldsymbol{\beta}\|_\infty / \|\boldsymbol{\beta}\|_2$, both null models concentrate rank mass near one, indicating samples that are more diffuse than the true patch-structured $\boldsymbol{\beta}$. The VAE failure on $\|\boldsymbol{\beta}\|_2$ matches the dark cortex map in Figure~\ref{fig:EEG_cortex}: a model that systematically samples small $\boldsymbol{\beta}$ values produces both small $\|\boldsymbol{\beta}\|_2$ relative to truth and small per-vertex source-space variance. The DDPM failure, by contrast, is invisible in the cortex map --- the spatial pattern is plausible --- but clearly visible in the SBC histograms, demonstrating that SBC catches calibration failures that qualitative diagnostic alone cannot. Crucially, the two null models achieve nearly identical reconstruction quality when paired with the same range model. This instantiates the cancellation pathology anticipated by the decomposition of Section~\ref{sec:uncertainty_decomp}: miscalibration of the intrinsic ambiguity distribution can persist even when end-to-end reconstruction is comparable, and the cascade structure is precisely what makes this miscalibration visible.

\section{Discussion}

Our experiments demonstrate that evaluating models merely by reconstruction quality fails to recognize important uncertainty miscalibration. In the EEG experiment, cascades with nearly identical reconstruction quality exhibited different uncertainty calibration. The cascade formulation addresses this by turning intrinsic ambiguity into an evaluable target. Its modular structure further allows each stage to be assessed independently: the range model through measurement consistency, and the null model through uncertainty calibration. The range--null combination results in Appendix~\ref{sec:additional_res} support this separation, showing that reconstruction quality varies primarily with the range model, whereas null-model choice primarily changes calibration behavior. Although the cascade imposes a particular factorization for evaluation, it does not constrain the model class used in either stage. Architectural advances in posterior sampling, diffusion models, or domain-specific networks can be incorporated as range or null models. 

We recognize the following limitations in our work. First, direct calibration of the intrinsic ambiguity distribution requires paired $(\mathbf{x}, \mathbf{y})$ data, so the diagnostic applies most directly to supervised settings rather than inference-only plug-and-play settings. Second, SBC provides a necessary but not sufficient check of calibration. Its sensitivity depends on the scalar statistics, which should be designed for the downstream application. Third, our current decomposition is most direct for known linear forward operators, where the geometry of the forward operator $\mathbf{A}$ gives an explicit range--null split. Extending the diagnostic to nonlinear or operator-uncertain inverse problems would require replacing this closed-form split with an approximate or learned notion of identifiable and ambiguous components. Beyond these limitations, an interesting direction is to investigate whether the cascade structure can be extended to amortized or foundation-model settings, where paired supervision may be partially available. Better-calibrated uncertainty estimates also have potential societal relevance in high-stakes inverse problems such as medical imaging. Distinguishing irreducible ambiguity from reducible inference uncertainty could support more reliable uncertainty reporting and guide data acquisition, although clinical or scientific deployment would require task-specific validation and domain oversight.

\clearpage
\enlargethispage{2\baselineskip} 
{
\small
\bibliographystyle{unsrtnat}
\bibliography{reference}
}







\appendix
\input{appendix}


\newpage

\end{document}

%% file: appendix.tex
\section{Graphical Model Interpretation}\label{app:pgm}
We introduce the graphical model to understand how the projection geometry naturally induces the decomposition of denoising inference and null space learning tasks, leading to the cascade model.

\begin{figure}[h]
    \center
    \includegraphics[width=0.6\columnwidth]{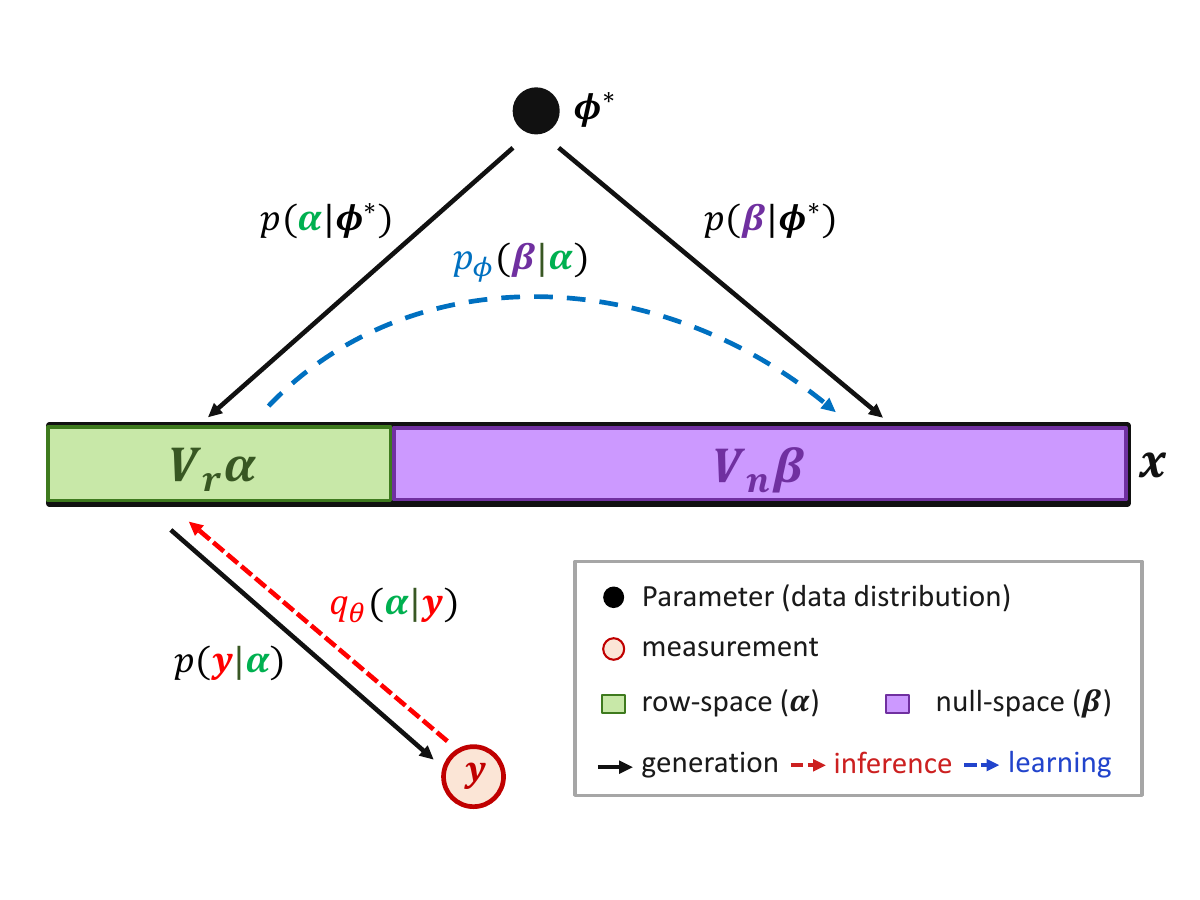}
    \caption{Probabilistic graphical model for the structured inverse problem under the realizability assumption $p_{\phi^*}(\mathbf{x}) = p_{\text{data}}(\mathbf{x})$. The latent variable $\mathbf{x}$ decomposes into a row-space component $\boldsymbol{\alpha} = \mathbf{V}_r^\top \mathbf{x}$ and a null-space component $\boldsymbol{\beta} = \mathbf{V}_n^\top \mathbf{x}$, reconstructed as $\mathbf{x} = \mathbf{V}_r\boldsymbol{\alpha} + \mathbf{V}_n\boldsymbol{\beta}$.
    The observation $\mathbf{y}$ informs $\boldsymbol{\alpha}$ via the posterior $q_\theta(\boldsymbol{\alpha} \mid \mathbf{y})$,
    approximating the true posterior $p(\boldsymbol{\alpha} \mid \mathbf{y})$, while $\boldsymbol{\beta}$ is drawn from the conditional $p_{\phi^*}(\boldsymbol{\beta} \mid \boldsymbol{\alpha})$.}
\label{fig:pgm}
\end{figure}

Figure~\ref{fig:pgm} illustrates this factorization as a probabilistic graphical model.
We assume the data $\mathbf{x}$ is generated under data distribution $p_{\text{data}}(\mathbf{x})$,
and suppose we are under the realizability assumption; we can parameterize it as $p_{\phi^*}(\mathbf{x})$. The data distribution $p_{\phi^*}(\mathbf{x})$ jointly generates the row-space and null-space components via marginals $p_{\phi^*}(\boldsymbol{\alpha})$ and $p_{\phi^*}(\boldsymbol{\beta})$. The cascade model we propose, Figure~\ref{fig:cascade_posterior}, factorizes the posterior over $\mathbf{x}$ into two sequential stages. First, the row-space component $\boldsymbol{\alpha}$ is inferred from the observation $\mathbf{y}$ via the posterior $q_\theta(\boldsymbol{\alpha} \mid \mathbf{y})$, which approximates the true posterior $p(\boldsymbol{\alpha} \mid \mathbf{y})$. Second, given the fixed row-space component $\boldsymbol{\alpha}^*$ — obtained directly from the range-space of $\mathbf{x}$ projected by $\mathbf{A}$ — the null-space component $\boldsymbol{\beta}$ is drawn from the conditional $p_{\phi^*}(\boldsymbol{\beta} \mid\boldsymbol{\alpha}^*)$. This conditional is learned via a generative model, which captures the statistical dependence between the null-space and the fixed row-space component. Since $\boldsymbol{\beta}$ lies entirely in the null-space of $\mathbf{A}$, it receives no direct signal from $\mathbf{y}$; its uncertainty is instead governed purely by the
learned prior $p_{\phi^*}(\boldsymbol{\beta} \mid \boldsymbol{\alpha}^*)$. This cascade structure reflects a fundamental asymmetry: $\boldsymbol{\alpha}$ lives in the row-space of $\mathbf{A}$ and is therefore constrained by the measurement, while $\boldsymbol{\beta}$ lives in the null-space of $\mathbf{A}$ and remains unidentifiable from $\mathbf{y}$ alone. The full reconstruction is then given
by $\mathbf{x} = \mathbf{V}_r\boldsymbol{\alpha} + \mathbf{V}_n\boldsymbol{\beta}$.

\section{Training and Architecture}
\begin{figure}[h]
\centering
\includegraphics[width=0.85\linewidth]{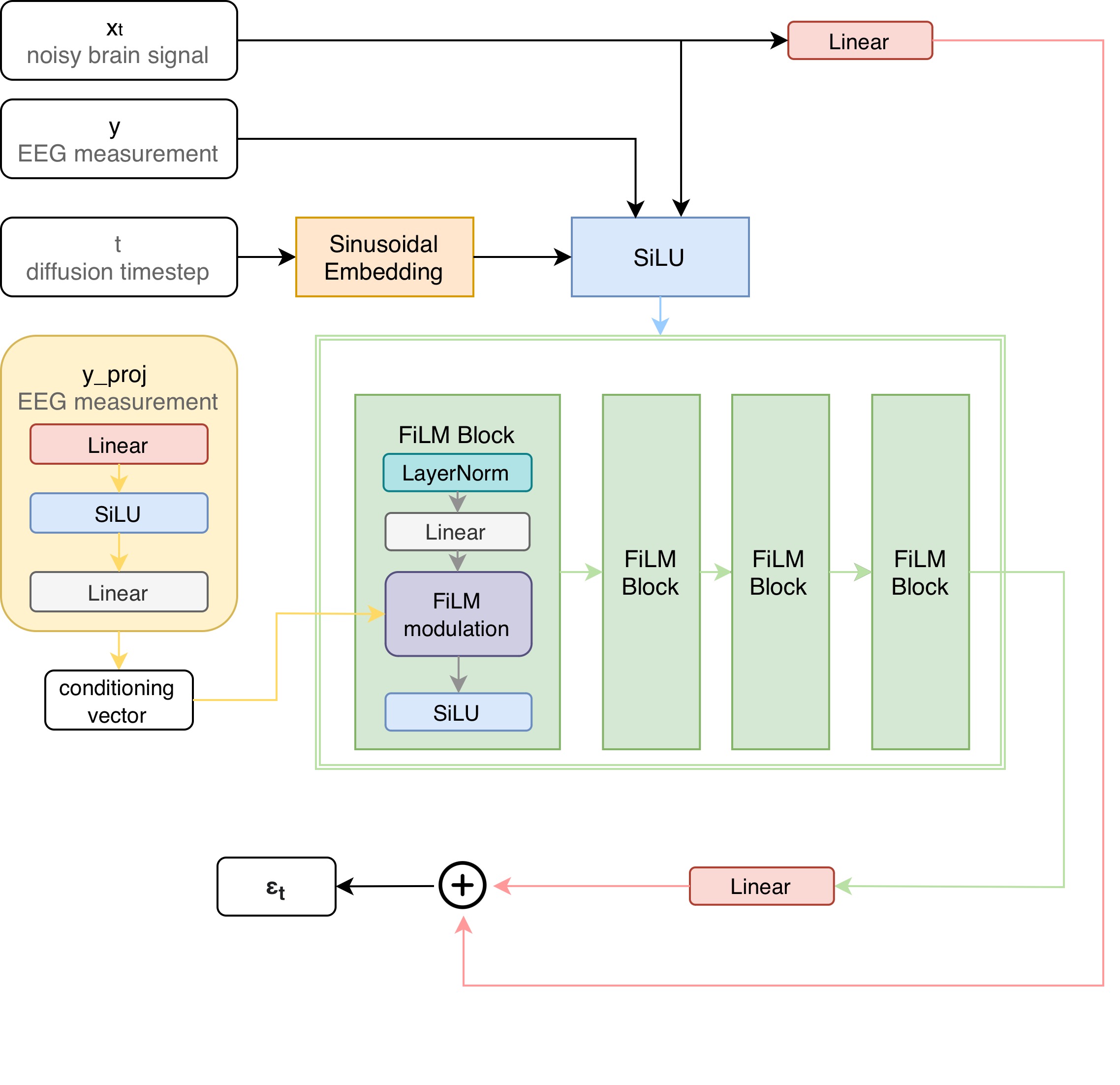}
\caption{Denoiser architecture used in the conditional diffusion posterior. The network takes the noisy source representation $\mathbf{x}_t$, EEG measurement $\mathbf{y}$, and diffusion timestep $t$ as input, and predicts the diffusion noise $\hat{\boldsymbol{\epsilon}}_\phi(\mathbf{x}_t,\mathbf{y},t)$. 
The EEG measurement is also encoded into a conditioning vector used for feature-wise linear modulation (FiLM) in each residual block. A linear skip connection from $\mathbf{x}_t$ is added to the final output.}
\label{fig:denoiser}
\end{figure}

We instantiate the cascade with a deterministic range model and a conditional DDPM null model, as an example. Other choices (probabilistic range models or VAE null models) are compatible with the same framework; Section~\ref{sec:additional_res} reports results for several such combinations. Algorithm~\ref{alg:cascade} summarizes training and inference for this instantiation, and Figure~\ref{fig:denoiser} illustrates the conditional DDPM denoiser architecture for the EEG experiment.

\begin{algorithm}[H]
\caption{Training and inference for the deterministic-range, DDPM-null cascade}
\label{alg:cascade}
\begin{algorithmic}[1]
\REQUIRE Training set $\{(\mathbf{x}_i, \mathbf{y}_i)\}_{i=1}^N$, forward operator $\mathbf{A}$

\STATE \textbf{Setup:} Compute SVD $\mathbf{A} = \mathbf{U}\boldsymbol{\Sigma}\mathbf{V}^\top$; partition $\mathbf{V} = [\mathbf{V}_r, \mathbf{V}_n]$
\STATE \textbf{Project:} $\boldsymbol{\alpha}_i^* \leftarrow \mathbf{V}_r^\top \mathbf{x}_i$, $\boldsymbol{\beta}_i^* \leftarrow \mathbf{V}_n^\top \mathbf{x}_i$ for all $i$
\STATE

\STATE \textit{// Train deterministic range model $f_{\boldsymbol{\theta}}: \mathbf{y} \mapsto \hat{\boldsymbol{\alpha}}$}
\FOR{epoch $= 1, \ldots, E_{\mathrm{range}}$}
    \FOR{minibatch $\{(\mathbf{y}_i, \boldsymbol{\alpha}_i^*)\}$}
        \STATE Update $\boldsymbol{\theta}$ to minimize $\|\boldsymbol{\alpha}_i^* - f_{\boldsymbol{\theta}}(\mathbf{y}_i)\|_2^2$
    \ENDFOR
\ENDFOR
\STATE

\STATE \textit{// Train null model with DDPM denoising loss}
\FOR{epoch $= 1, \ldots, E_{\mathrm{null}}$}
    \FOR{minibatch $\{(\boldsymbol{\alpha}_i^*, \boldsymbol{\beta}_i^*)\}$}
        \STATE Sample $t \sim \mathrm{Uniform}\{1,\ldots,T\}$, $\boldsymbol{\epsilon} \sim \mathcal{N}(\mathbf{0}, \mathbf{I})$
        \STATE $\boldsymbol{\beta}_t \leftarrow \sqrt{\bar{\alpha}_t}\,\boldsymbol{\beta}_i^* + \sqrt{1-\bar{\alpha}_t}\,\boldsymbol{\epsilon}$
        \STATE Update $\boldsymbol{\phi}$ to minimize $\|\boldsymbol{\epsilon} - \boldsymbol{\epsilon}_{\boldsymbol{\phi}}(\boldsymbol{\beta}_t, \boldsymbol{\alpha}_i^*, t)\|_2^2$
    \ENDFOR
\ENDFOR
\STATE

\STATE \textit{// Inference: sample $\hat{\mathbf{x}} \sim p(\mathbf{x} \mid \mathbf{y})$}
\STATE Compute $\hat{\boldsymbol{\alpha}} = f_{\boldsymbol{\theta}}(\mathbf{y})$
\STATE Sample $\hat{\boldsymbol{\beta}} \sim p_{\boldsymbol{\phi}}^{\mathrm{null}}(\boldsymbol{\beta} \mid \hat{\boldsymbol{\alpha}})$ via reverse DDPM diffusion
\STATE \textbf{return} $\hat{\mathbf{x}} = \mathbf{V}_r \hat{\boldsymbol{\alpha}} + \mathbf{V}_n \hat{\boldsymbol{\beta}}$
\end{algorithmic}
\end{algorithm}

\clearpage
\section{Gaussian Analytical Example: Data Generation and Model Details}
\label{app:gaussian_details}

\paragraph{Generative model.}
Data are generated as a Gaussian instantiation of the factorization in
Eq.~\eqref{eq:factorization}, where $\boldsymbol{y}$ depends only on
$\boldsymbol{\alpha}$ by construction, so that
$\boldsymbol{\beta} \perp \boldsymbol{y} \mid \boldsymbol{\alpha}$ holds exactly:
\begin{align}
\boldsymbol{\alpha} &\sim \mathcal{N}(\mathbf{0}, \mathbf{I}_r), \\
\boldsymbol{\beta} \mid \boldsymbol{\alpha} &\sim
    \mathcal{N}(\mathbf{C}\boldsymbol{\alpha},\, \boldsymbol{\Sigma}_{\boldsymbol{\eta}}), \\
\boldsymbol{y} \mid \boldsymbol{\alpha} &\sim
    \mathcal{N}(\mathbf{A}\boldsymbol{\alpha},\, \sigma_y^2\mathbf{I}_n).
\end{align}
We use dimensions $r = 32$, $q = 64$, $n = 32$, and measurement noise
$\sigma_y = 0.3$.

\paragraph{Parameter construction.}
The forward operator $\mathbf{A} \in \mathbb{R}^{n \times r}$ is drawn as a
random Gaussian matrix and orthogonalized via QR decomposition, so that
$\mathbf{A}^\top\mathbf{A} = \mathbf{I}_r$ and the identifiable subspace is
isotropically covered.
The cross-covariance matrix $\mathbf{C} \in \mathbb{R}^{q \times r}$ is drawn
i.i.d.\ from $\mathcal{N}(0, r^{-1})$, so that
$\|\mathbf{C}\boldsymbol{\alpha}\|^2 \approx \|\boldsymbol{\alpha}\|^2$ in
expectation.

The intrinsic ambiguity covariance $\boldsymbol{\Sigma}_{\boldsymbol{\eta}} \in
\mathbb{R}^{q \times q}$ is constructed as $\boldsymbol{\Sigma}_{\boldsymbol{\eta}}
= \mathbf{Q}\boldsymbol{\Lambda}\mathbf{Q}^\top$, where $\mathbf{Q} \in O(q)$
is a random orthogonal matrix obtained by QR decomposition of a Gaussian matrix,
and $\boldsymbol{\Lambda} = \mathrm{diag}(\lambda_1, \ldots, \lambda_q)$ has
eigenvalues spaced geometrically from $\lambda_{\max} = 8.0$ to
$\lambda_{\min} = 0.1$ (condition number $\approx 80$).
This construction yields a fully correlated covariance with off-diagonal
structure, ensuring that calibration requires matching not only marginal scales
but also the geometry of the ambiguous subspace.
Samples $\boldsymbol{\eta} \sim \mathcal{N}(\mathbf{0},
\boldsymbol{\Sigma}_{\boldsymbol{\eta}})$ are drawn via the Cholesky factor
$\mathbf{L}$ as $\boldsymbol{\eta} = \mathbf{L}\mathbf{z}$,
$\mathbf{z} \sim \mathcal{N}(\mathbf{0}, \mathbf{I}_q)$.

\paragraph{Analytical posteriors.}
The Gaussian structure admits closed-form posteriors used as ground truth for
calibration:
\begin{align}
\boldsymbol{\Sigma}_{\boldsymbol{\alpha}|y}
    &= \bigl(\mathbf{I}_r + \sigma_y^{-2}\mathbf{A}^\top\mathbf{A}\bigr)^{-1}, \quad
\boldsymbol{\mu}_{\boldsymbol{\alpha}|y}
    = \sigma_y^{-2}\boldsymbol{\Sigma}_{\boldsymbol{\alpha}|y}\mathbf{A}^\top\mathbf{y}, \\
\mathrm{Cov}(\boldsymbol{\beta} \mid \boldsymbol{y})
    &= \mathbf{C}\,\boldsymbol{\Sigma}_{\boldsymbol{\alpha}|y}\,\mathbf{C}^\top
       + \boldsymbol{\Sigma}_{\boldsymbol{\eta}}, \\
\boldsymbol{\beta} \mid \boldsymbol{\alpha}^*
    &\sim \mathcal{N}\!\left(\mathbf{C}\boldsymbol{\alpha}^*,\,
       \boldsymbol{\Sigma}_{\boldsymbol{\eta}}\right).
\end{align}
The oracle ambiguity distribution $p(\boldsymbol{\beta} \mid \boldsymbol{\alpha}^*)$
is thus available in closed form, enabling direct comparison against the learned
null model.

\paragraph{Null model training and evaluation.}
A conditional DDPM is trained on $N = 100{,}000$ i.i.d.\ samples of
$(\boldsymbol{\alpha}^*, \boldsymbol{\beta}^*)$ pairs using a linear noise
schedule with $T = 1000$ diffusion steps ($\beta_1 = 10^{-4}$,
$\beta_T = 0.02$), a two-block MLP denoiser with hidden dimension 256, the
Adam optimizer at learning rate $3 \times 10^{-4}$, and batch size 256, for
$10^5$ gradient steps.
SBC diagnostics are computed using $M_{sbc} = 200$ posterior samples per test point over $N_{\mathrm{sbc}} = 500$ held-out values of $\boldsymbol{\alpha}^*$, using the scalar test statistics
$T_1 = \|\boldsymbol{\beta}\|_2$ and $T_2 = \|\boldsymbol{\beta}\|_\infty / \|\boldsymbol{\beta}\|_2$.

\section{EEG cascade results}\label{sec:additional_res}

\begin{figure}[h]
\centering
\includegraphics[width=\linewidth]{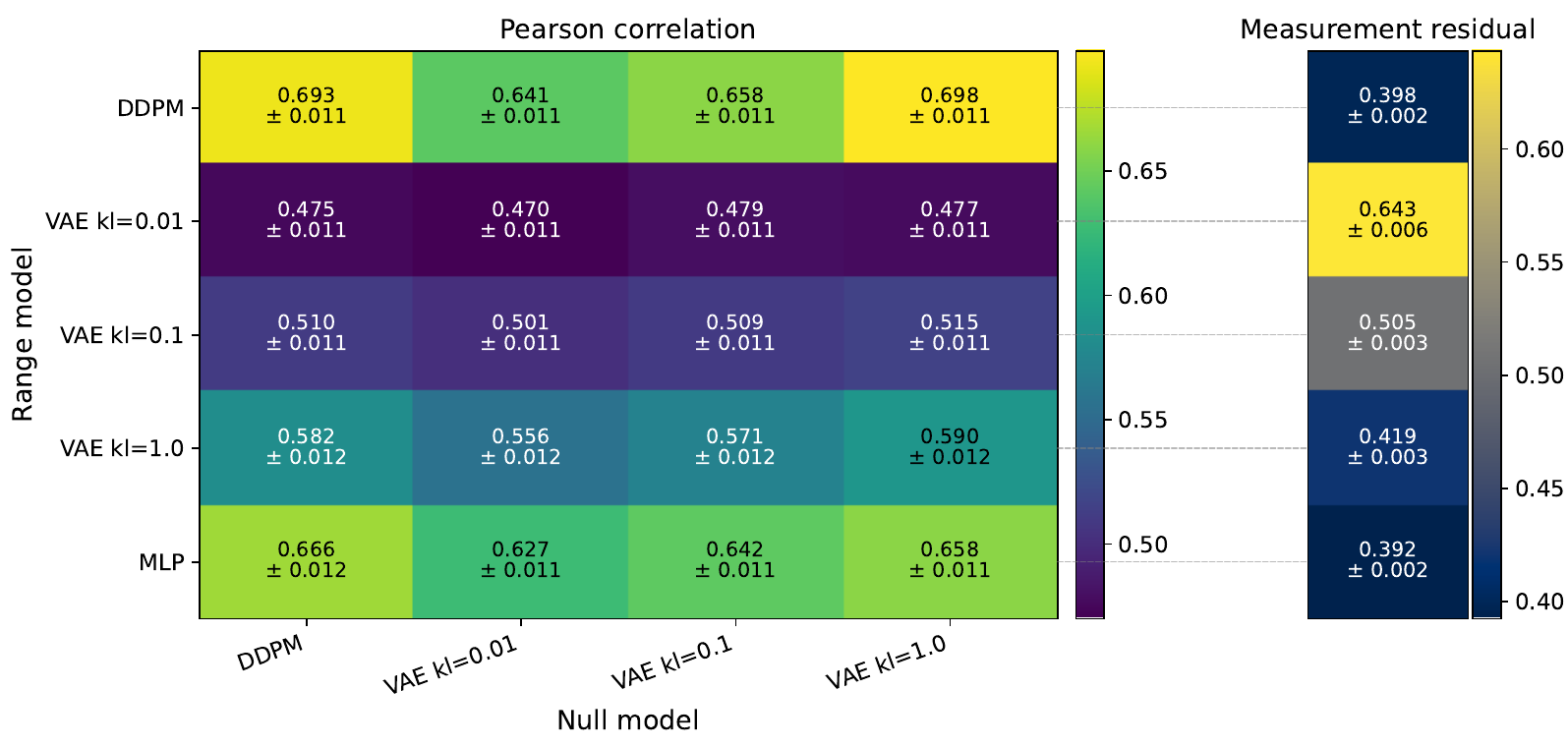}
\caption{Pearson correlation between the posterior mean reconstruction and the ground-truth source signal across range--null model assemblies, averaged over 450 held-out test cases. Cells report mean $\pm$ s.e.m. across test cases. The right column reports the measurement residual, $\|\mathbf{y} - \mathbf{A}\hat{\mathbf{x}}\|_2 / \|\mathbf{y}\|_2$, which depends only on the range-stage reconstruction.}
\label{fig:EEG_res_grid}
\end{figure}

Figure~\ref{fig:EEG_res_grid} reports posterior mean reconstruction quality for all trained range--null combinations. The dominant pattern is that Pearson correlation varies primarily across range models, while differences across null models are comparatively small within each row. This is expected because the posterior mean reconstruction is strongly influenced by the measurement-identifiable component, which is determined by the range model. In contrast, the null model primarily controls the distribution of intrinsically ambiguous components.

This grid supports the model-selection motivation of the main text. A practitioner selecting a cascade using reconstruction correlation alone would find several near-equivalent combinations. However, as shown in Section~\ref{sec:eeg}, models with similar reconstruction quality can differ substantially in their calibration of intrinsic ambiguity. Thus, the range--null grid illustrates why reconstruction metrics alone are insufficient for selecting posterior models in ill-posed inverse problems.

\newpage
\section{Noise propagation bound for accelerated MRI}
\label{app:mri_bound}

We provide a short argument for why the propagated uncertainty term in Eq.~\ref{eq:total_var} is controlled by measurement noise in the accelerated MRI setting. 

The propagated uncertainty term in Eq.~\ref{eq:total_var} is $\mathbb{V}_{\boldsymbol{\alpha} \mid \mathbf{y}}\!\left(\mathbb{E}[\boldsymbol{\beta} \mid \boldsymbol{\alpha}]\right)$. 
In the MRI notation, this corresponds to $\mathbb{V}_{\hat{\mathbf{x}}_{\boldsymbol{\alpha}}}\!\left(m(\hat{\mathbf{x}}_{\boldsymbol{\alpha}})\right)$, where $m(\mathbf{x}_{\boldsymbol{\alpha}}) = \mathbb{E}_{p_\phi^\mathrm{null}}[\mathbf{x}_{\boldsymbol{\beta}} \mid \mathbf{x}_{\boldsymbol{\alpha}}]$ is the conditional null mean.
The forward operator is $\mathbf{A=MF}$, where $\mathbf{F}$ is the two-dimensional Fourier transform and $\mathbf{M}$ is a binary k-space mask. For a clean image $\mathbf{x}$, the oracle identifiable component is $\mathbf{x}_{\boldsymbol{\alpha}}^* = \mathbf{F}^{-1}\mathbf{MFx}$. Under noisy measurements $\mathbf{y}_\sigma = \mathbf{MFx} + \boldsymbol{\varepsilon}_\sigma$, the closed-form identifiable estimate is
\[
\hat{\mathbf{x}}_{\boldsymbol{\alpha}} = \mathbf{F}^{-1}\mathbf{y}_\sigma = \mathbf{x}_{\boldsymbol{\alpha}}^* + \mathbf{F}^{-1}\boldsymbol{\varepsilon}_\sigma .
\]

Thus $\hat{\mathbf{x}}_{\boldsymbol{\alpha}}-\mathbf{x}_\alpha^* 
= \mathbf{F}^{-1}\boldsymbol{\varepsilon}_\sigma$.
For a unitary normalization of the Fourier transform, $\|\mathbf{F}^{-1}\boldsymbol{\varepsilon}_\sigma\|_2^2 = \|\boldsymbol{\varepsilon}_\sigma\|_2^2$, and therefore $\mathbb{E}\|\hat{\mathbf{x}}_\alpha - \mathbf{x}_\alpha^*\|_2^2 = O(\sigma^2)$.

Under the assumption that $m$ is $L$-Lipschitz, then
\begin{align}
\text{tr}(\mathbb{V}_{\hat{\mathbf{x}}_\alpha}) \! 
\left(m(\hat{\mathbf{x}}_\alpha)\right)
&\leq
\mathbb{E}\left\|m(\hat{\mathbf{x}}_\alpha)-m(\mathbf{x}_\alpha^*)\right\|_2^2 \nonumber\\
&\leq
L^2\mathbb{E}\|\hat{\mathbf{x}}_\alpha-\mathbf{x}_\alpha^*\|_2^2 = O(\sigma^2).
\end{align}
Hence, the propagated uncertainty term vanishes as $\sigma\to 0$, supporting the interpretation of Fig.~\ref{fig:MRI_decomp}(a): intrinsic ambiguity is fixed when conditioning on $\mathbf{x}_{\boldsymbol{\alpha}}^*$, while additional null uncertainty under noisy conditioning is controlled by perturbations in $\hat{\mathbf{x}}_{\boldsymbol{\alpha}}$.

\newpage
\section{Simulation-Based Calibration for High-Dimensional Parameters}
\label{app:sbc-high-dim}

Simulation-Based Calibration (SBC) \citep{talts2018validating} provides a principled diagnostic
for posterior inference by exploiting a self-consistency property of Bayes' rule. If
$\theta^{\mathrm{sim}} \sim p(\theta)$ and $y^{\mathrm{sim}} \sim p(y \mid \theta^{\mathrm{sim}})$,
then $\theta^{\mathrm{sim}}$ is itself a draw from the posterior $p(\theta \mid y^{\mathrm{sim}})$,
and its rank among approximate posterior draws $\{\theta^{(m)}\}_{m=1}^M$ must be uniformly
distributed on $\{0, \ldots, M\}$. Concretely, the rank statistic is
\begin{equation}
    r_k = \sum_{m=1}^{M} \mathbf{1}\!\left[\theta_k^{(m)} < \theta_k^{\mathrm{sim}}\right],
\end{equation}
and any systematic deviation from uniformity in $\{r_k\}$ signals that the approximate posterior
is miscalibrated. This diagnostic is sensitive to three distinct failure modes: model
misspecification, sampler pathology, and implementation errors.

\paragraph{Challenges in High Dimensions.}
For $\boldsymbol{\theta} \in \mathbb{R}^D$, the standard procedure yields $D$ separate rank sequences, one
per coordinate. This raises two difficulties. First, simultaneous uniformity tests across $D$
parameters require multiple-testing correction, which reduces statistical power. Second, and more
fundamentally, passing $D$ marginal tests does not imply joint calibration: a posterior can have
correct marginals while being badly wrong about inter-coordinate correlations.

\paragraph{Scalar Aggregates for Exchangeable Parameters.}
When the coordinates of $\boldsymbol{\theta}$ are semantically homogeneous and exchangeable under the prior
--- for instance, when each $\boldsymbol{\theta}_k$ represents the posterior mean of the $k$-th data dimension
--- scalar aggregates of $\boldsymbol{\theta}$ provide computationally efficient and interpretable SBC
diagnostics. We propose two complementary statistics.

The first is the $\ell_2$ norm $\|\boldsymbol{\theta}\|_2$, which aggregates total energy across dimensions.
Its rank probes whether the posterior captures the correct overall scale: systematic shrinkage or
overdispersion manifests as a rank distribution skewed toward $0$ or $M$, respectively.

The second is the peakedness ratio $\|\boldsymbol{\theta}\|_\infty / \|\boldsymbol{\theta}\|_2$, which measures the degree
of concentration of $\boldsymbol{\theta}$ across its coordinates. This ratio lies in $[D^{-1/2}, 1]$, with
values near $D^{-1/2}$ indicating diffuse, roughly uniform energy and values near $1$ indicating
near-sparse concentration in a single coordinate. A posterior that over-smooths a sparse signal
will produce a ratio systematically smaller than the truth, and SBC will detect this as a
rank distribution biased toward $0$.

\paragraph{Complementarity.}
These two statistics probe orthogonal aspects of the posterior geometry, as summarized below:
\begin{center}
\begin{tabular}{lll}
\toprule
\textbf{Statistic} & \textbf{Sensitive to} & \textbf{Blind to} \\
\midrule
$\|\boldsymbol{\theta}\|_2$ & Overall scale / shrinkage & Sparsity structure \\
$\|\boldsymbol{\theta}\|_\infty / \|\boldsymbol{\theta}\|_2$ & Sparsity / concentration & Overall energy level \\
\bottomrule
\end{tabular}
\end{center}
Running SBC on both simultaneously provides a two-dimensional diagnostic well-suited to
structured, high-dimensional posteriors where sparsity is a meaningful property of the signal.